\def\year{2018}\relax
\documentclass[letterpaper]{article} %DO NOT CHANGE THIS
\usepackage{aaai18}  %Required
\usepackage{times}  %Required
\usepackage{helvet}  %Required
\usepackage{courier}  %Required
\usepackage{url}  %Required
\usepackage{graphicx}  %Required

\usepackage{epsfig}
\usepackage{amssymb}
\usepackage{multirow}
\usepackage{bbding}
\usepackage{xspace}
\usepackage{enumerate}
\usepackage{makecell}
\usepackage{caption}
\usepackage{subcaption}

\usepackage{algorithm}
\usepackage{algorithmic}
\usepackage{mathrsfs} % For LaTeX2e
\usepackage{amsmath,bm}
\usepackage{color}
\usepackage{caption}
\usepackage{booktabs}
\usepackage{array}
\usepackage{float}
\def\eg{{\em e.g.}}
\def\ie{{\em i.e.}}
\def\etal{{\em et al.}}

\begin{document}
% The file aaai.sty is the style file for AAAI Press
% proceedings, working notes, and technical reports.
%
\title{Facial Landmarks Detection by Self-Iterative Regression based Landmarks-Attention Network}
\author{Tao Hu$^1$, Honggang Qi$^1$, Jizheng Xu$^2$, Qingming Huang$^1$\\
$^1$ University of Chinese Academy of Sciences, Beijing, China\\
$^2$ Microsoft Research Asia, Beijing, China\\
\textit {hutao16@mails.ucas.ac.cn, hgqi@ucas.ac.cn}
}
\maketitle
%%%%%%%%%%%%%%%%%%%%%%%%%%%%%%%%%%%%%%%%%%%%%%%%%%%%%%%%%%%%%%%%%%%%%
%%%%%%%%%%%%%%%%%%%%%%%%%%%%%%%%%%%%%%%%%%%%%%%%%%%%%%%%%%%%%%%%%%%%%
\begin{abstract}
Cascaded Regression (CR) based methods have been proposed to solve facial landmarks detection problem, which learn a series of descent directions by multiple cascaded regressors separately trained in coarse and fine stages. They outperform the traditional gradient descent based methods in both accuracy and running speed. However, cascaded regression is not robust enough because each regressor's training data comes from the output of previous regressor. Moreover, training multiple regressors requires lots of computing resources, especially for deep learning based methods. In this paper, we develop a Self-Iterative Regression (SIR) framework to improve the model efficiency. Only one self-iterative regressor is trained to learn the descent directions for samples from coarse stages to fine stages, and parameters are iteratively updated by the same regressor. Specifically, we proposed Landmarks-Attention Network (LAN) as our regressor, which concurrently learns features around each landmark and obtains the holistic location increment. By doing so, not only the rest of regressors are removed to simplify the training process, but the number of model parameters is significantly decreased. The experiments demonstrate that with only 3.72M model parameters, our proposed method achieves the state-of-the-art performance.
\end{abstract}

%%%%%%%%%%%%%%%%%%%%%%%%%%%%%%%%%%%%%%%%%%%%%%%%%%%%%%%%%%%%%%%%%%%%%
\section{Introduction} \label{sec:intro}
Facial landmarks detection is one of the most important techniques in face analysis, such as face recognition, facial animation and 3D face reconstruction. It aims to detect the facial landmarks such as eyes, nose and mouth, namely predicting the location parameters of landmarks. 

Researchers usually regard this task as a typical non-linear least squares problem~\cite{Xiong:SDM:CVPR2013}. The Newton's method and its variants are the traditional gradient based solution, whose convergence rate is quadratic and is guaranteed to converge, provided that the initial estimate is sufficiently close to the minimum. However, when the objective function is not differentiable(\eg~SIFT\cite{SIFT:ijcv:Lowe04}) or the Hessian matrix is not positive definite, the method won't works well~\cite{Xiong:SDM:CVPR2013,Xiong:GSDM:CVPR2015}.
% \begin{figure}[t]
% \centering
% \includegraphics[width=1.0\linewidth]{Figs/SIR_VS_CR.png}
% \caption{\small Facial landmarks detection process of Cascaded Regression(a) and Self-Iterative Regression(b). To predict the landmarks' location parameters, the CR based methods require multiple regressors, while SIR only trains one regressor and updates parameters iteratively.}
% \label{fig:SIR_VS_CR}
% \end{figure}
\begin{figure}[t]
\begin{subfigure}{.25\textwidth}
  \centering
  \includegraphics[width=\linewidth]{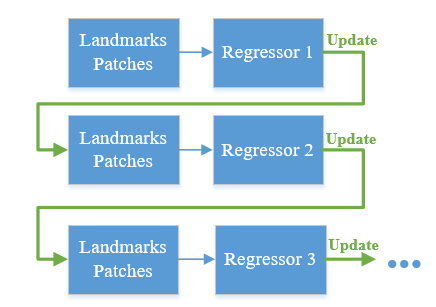}
  \caption{Cascaded Regression.}
  \label{fig:sfig1}
\end{subfigure}%
\begin{subfigure}{.25\textwidth}
  \centering
  \includegraphics[width=\linewidth]{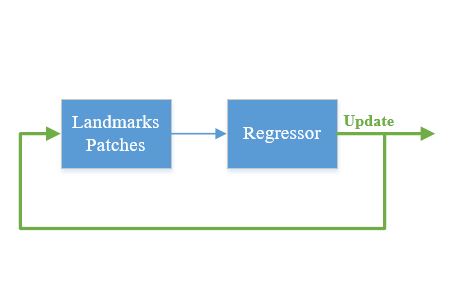}
  \caption{Self-Iterative Regression.}
  \label{fig:sfig2}
\end{subfigure}
\caption{Facial landmarks detection process of Cascaded Regression(a) and Self-Iterative Regression(b). To predict the landmarks' location parameters, the CR based methods require multiple regressors, while SIR just need one regressor and updates parameters iteratively.}
\label{fig:SIR_VS_CR}
\end{figure}

In recent years, cascaded regression based methods~\cite{Dollar:CPR:CVPR2010,Cao:ESR:CVPR2012,Xiong:SDM:CVPR2013,ShaoqinRen:LBF:CVPR2014,Xiangyu:3DDFA:CVPR2016,FengLiu:JFAFR:ECCV2016,Tzimiropoulos:PO-CR:CVP2015,Zhuowen:cvpr08} have been proposed and applied to solve the non-linear least squares problem. They usually train multiple regressors to predict the parameters' increment sequentially, which outperform the traditional gradient descent based methods in both accuracy and running speed. Moreover, deep learning based cascaded regression methods~\cite{SunWT:cascaded-cnn:cvpr13,Zhang:MTL:ECCV2014,Trigeorgis:MDM:CVPR16,XiaoFXLYK:RAR:ECCV2016,eccv14:ZhangSKC:CFAN} are widely leveraged for this task because of the powerful ability to extract the discriminative feature. However, when applying cascaded regression system, three main problems arise: (1) Each regressor just works well in its local data space, when previous regressor predicts the false descent direction, the final results are very likely to drift away; (2) In general, higher accuracy can be obtained by adding more cascaded regressors, while it will increase model storage memory and computing resources; (3) Subsequent regressors usually cannot be activated for training until previous regressors finished their training process, which increases the system complexity.

In this paper, we develop a Self-Iterative Regression (SIR) framework to solve the above issues. By means of the powerful representation of Convolutional Neural Network (CNN), we only train \textit{one} regressor to learn the descent directions in coarse and fine stages together. The training data is obtained by random sampling in the parameter space, and in the testing process, parameters are updated iteratively by calling the same regressor, which is dubbed Self-Iterative Regression. The testing process is illustrated in Figure~\ref{fig:SIR_VS_CR}(b). The experimental results show that for deep learning based method, one regressor achieves comparable performance to state-of-the-art multiple cascaded regressors and significantly reduce the training complexity. Moreover, to obtain discriminative landmarks features, we proposed a Landmarks-Attention Network (LAN), which focuses on the appearance around landmarks. It first concurrently extracts local landmarks' features and then obtains the holistic increment, which significantly reduces the dimension of the final feature layer and the number of model parameters. The contributions of this paper are summarized as follows:
\begin{enumerate}
  \item We propose a novel regression framework called SIR to solve the non-linear least squares problem, which simplifies the cascaded regression framework and obtains state-of-the-art performance in facial landmarks detection task.
  \item The Landmarks-Attention Network (LAN) is developed to independently learn discriminative features around each landmarks, which significantly reduces the dimension of feature layer and the number of model parameters.
  \item Experimental results on several publicly available benchmarks demonstrate the effectiveness of the proposed method.
\end{enumerate}

\section{Related Work}\label{previous_works}
In this section, we will review related works in solving non-linear least squares problems, especially facial landmarks detection problem.

{\noindent \textbf{Cascaded Regression based Methods.}} Cascaded regression was first introduced by Doll\'{a}r~\etal~\cite{Dollar:CPR:CVPR2010}, which trains a fixed cascaded linear week regressors to predict the pose parameter of the object. Then, Xiong~\etal~describes the cascaded regression problem as a general learning framework called Supervised Descent Method (SDM) in~\cite{Xiong:SDM:CVPR2013}. It avoids computing the Jacobian and Hessian matrix by learning a sequence of local descent directions to minimize the non-linear least squares function. To accelerate the running speed of facial landmarks detection, LBF is developed in~\cite{ShaoqinRen:LBF:CVPR2014}, which learns local binary feature with random forest\cite{Breiman01:random_forest} and obtains the final output by jointly learning the linear regression. To obtain a robust initialization, the CFSS~\cite{Zhu:CFSS:CVPR2015} first performs a coarse shape search over the shape spaces and then constrains the subsequent refinement regressors by the coarse shape.

{\noindent \textbf{Deep Learning based Methods.}} The CNN based methods can extract more discriminative features than above methods. Sun~\etal~\cite{SunWT:cascaded-cnn:cvpr13} presents a deep cascaded regression based method by cascading three levels of CNNs and it regress the location of facial landmarks with the coarse-to-fine strategy. The disadvantage is obvious: too many CNNs ($23$ CNNs in their work) need to be trained, which requires too much computing resources.

Zhang~\etal~develops a Coarse-to-Fine Auto-encoder Network (CFAN)~\cite{eccv14:ZhangSKC:CFAN}, which consists of multiple Stacked Auto-encoder Networks (SANs). The first SAN quickly predicts the preliminary location of landmarks by a low-resolution image, and the subsequent SANs then refine the location with higher and higher resolution.

Trigeorgis~\etal~proposed the Mnemonic Descent Method (MDM)~\cite{Trigeorgis:MDM:CVPR16}, which regards the non-linear least squares optimization as a dynamic process. The Recurrent Neural Network (RNN) is introduced to maintain an internal memory unit that accumulates the history information so as to relate the cascaded refinement process. 

Jo{\~{a}}o~\etal~proposed a iterative error feedback~\cite{CarreiraAFM:IEF:cvpr16} method to solve the human pose extimation problems. Same with MDM, their training data is generated by previous stages, while ours is obtained by random sampling in coarse stages and fine stages, which simplifies the training process.

Xiao~\etal~\cite{XiaoFXLYK:RAR:ECCV2016} propose a Long Short Term Memory (LSTM) based recurrent attentive-refinement network, which also follows the pipeline of cascaded regressions. Instead of updating all landmarks location together, it first extracts reliable landmarks by a CNN and then infers locations of the rest noisy landmarks, resulting in improved accuracy. However, these deep cascaded regression methods usually require more computing resources and also suffer from the same drawbacks as discussed above.

\section{Cascaded Regression}
Before introducing our method, we begin with the cascaded regression framework in brief for better understanding. As illustrated in Figure~\ref{fig:SIR_VS_CR}(a), in the training process of cascaded regression, $K$ regressors ($R_1, R_2,\cdots, R_K$) are trained sequentially. Each regressor $R_k$ is computed by minimizing the expected loss between the predicted and the optimal parameters's increment. It is formulated as
\begin{equation}
R_k = \underset{R_k}{\operatorname{argmin}}\sum_{i} \| \Delta\theta_{k,i}^*-R_k(x_{k,i}) \|^2_2, k=1,2,\cdots,K,
\label{equ:cascade_regression}
\end{equation}
where $x_{k, i}$ is $i_{th}$ example in $k_{th}$ regression process, $\Delta\theta_{k,i}^* = \theta_i^*- \theta_{k, i}$  is the corresponding target increment, \ie, the difference between ground truth parameter $\theta_i^*$ and present parameter $\theta_{k, i}$. After obtaining $R_k$, the target parameter is updated by Equ.~\eqref{equ:next_regression},
\begin{equation}
\theta_{k+1, i} = \theta_{k,i} + R_k(x_{k, i}).
\label{equ:next_regression}
\end{equation}

Then, new training dataset will be generated according to the updated parameter for the next regression ~\cite{Xiong:SDM:CVPR2013}.

In the testing process, parameter will be sequentially refined by these cascaded regressors in Equ~\ref{equ:next_regression}.

\section{Self-Iterative Regression}\label{framework}
\begin{figure*}[t]
\centering
\includegraphics[width=0.85\linewidth]{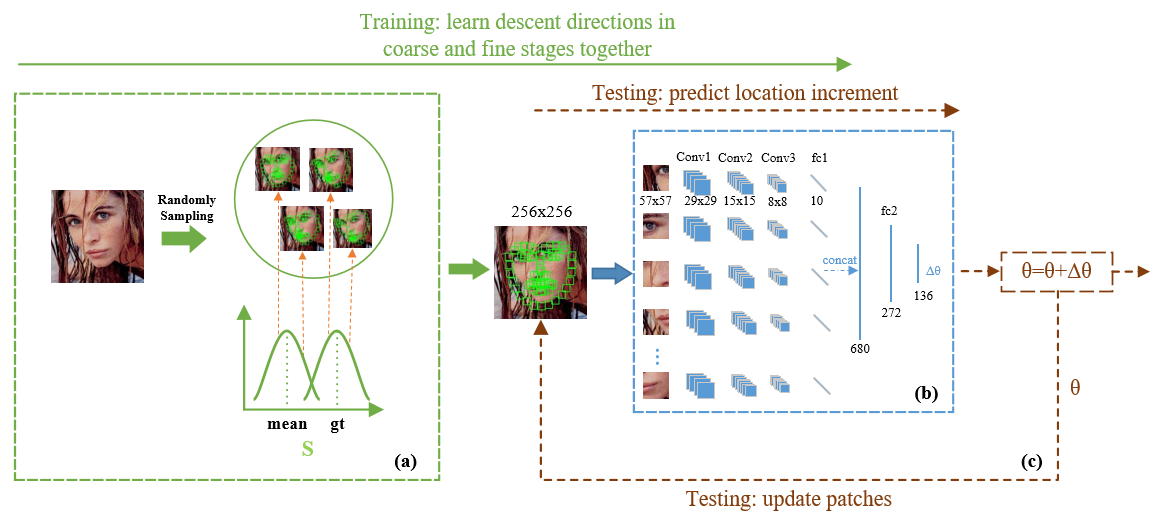}
\caption{Training and testing process of the proposed SIR. (a) random sampling process. (b) Landmarks-Attention Network. (c) Iterative predicting and updating process. The training process consists of (a) and (b), while the testing process consists of (b) and (c). In the figure, one of the dimension of facial Landmarks Model parameter $S$ is showed, and $\theta$ is landmarks' location parameter.}
\label{fig:Framework}
\end{figure*}

In this section, we will describe our facial landmarks detection method including the Gaussian random sampling and the Landmarks-Attention Network in detail. The overall procedure is presented in Figure~\ref{fig:Framework}.

\subsection{Gaussian random Sampling}
Generating training data is key important process in our method. Cascaded regression generates training data according to previous regressor, while our method obtains it by random sampling, which includes most possible landmarks distribution from coarse stages to fine stages. Let $(x_j, y_j)$ be the $j_{th}$ landmark's position coordinates and $\theta = (x_1, y_1, x_2, y_2, ..., x_M, y_M)$ be all landmarks' location parameters, where $M$ is the total number of facial landmarks. It is not a good choice to directly sample in location parameter $\theta$ since its dimension is so high that the training process will be hard to converge and is very likely to generate unreasonable face shape. To improve the effectiveness of sampling process, we indirectly obtain sampling location $\theta$ according to a new facial landmarks model that is similar to 3D Morphable Model (3DMM)~\cite{Volker:3DMM:SIGGRAPH99}. Facial landmarks distribution will be represented by pose and shape parameter.

{\noindent \textbf{Facial Landmarks Model.}} We obtain intrinsic face shape parameter by Principal Component Analysis (PCA) and pose parameter(including 2D translation, in-plane rotation and scale) by geometry transformation. The shape, translation, rotation angle and scale coefficient are represented by $\alpha$, $t_{2d}$, $\beta$, $f$ respectively. Finally, facial landmarks model parameters can be represented by $S = [\alpha, t_{2d}, \beta, f]$. $S$ and $\theta$ are two kinds of representation for facial landmarks. $S$ can be converted to $\theta$ by
\begin{equation}
\theta(S)=f*R_{\beta}*(S_0 + A*\alpha) + t_{2d},
\label{equ:face_model}
\end{equation}
where $S_0$ is the mean shape, $A$ is the PCA shape matrix and $R_{\beta}=\begin{bmatrix}\cos\beta&-\sin\beta\\ \sin\beta&\cos\beta \end{bmatrix}$ is the rotation matrix with angle $\beta$. 

Random sampling in facial landmarks model $S$ and then converting to location parameter $\theta$ makes the sampling process easier to control and generates more reasonable landmarks' distribution. 

{\noindent \textbf{Sampling space}}. For each face $I$, let $S_{\text{gt}}$ represent its ground truth facial landmarks model parameters. We random select values in each dimension of $S$ obeying distribution $D$ which is a union set of two Gaussian distribution. The sampling space of each face is represented by
\begin{equation}
\label{equ:sampling_space}
D \sim \{N(S_0, \sigma) \cup N(S_{gt}, \sigma) \},
\end{equation}
where $N(\cdot,\cdot)$ represents Gaussian distribution, and $\sigma$ is its standard deviation.

We adopt this sampling distribution because training regressor around mean location and ground truth location affects the performance in coarse and fine stages, respectively, and the final location error usually obeys Gaussian distribution. The value of standard deviation $\sigma$ affects the final performance. System with larger $\sigma$ will contain more training space which makes the system more robust, while the final accuracy may decrease because sampling probability around ground truth will decrease and vise versa. The effect of $\sigma$ will be discussed in the \emph{Experiments} section.

For $i$-th image in the $t$-th sampling period, sampling parameter $S_{t, i}$ is obtained by random selecting a value in Equ.~\eqref{equ:sampling_space}. We then calculate location parameters $\theta_{t,i}$ by Equ~\eqref{equ:face_model} and extract patches $P_{t,i}$ in location $\theta_{t,i}$. Finally, we set $P_{t,i}$ as the training input data and set $\Delta \theta_{t,i} = \theta_i^* - \theta_{t,i}$ as regressor's corresponding target increment. The process is also summarized in Algorithm~\ref{alg:sampling}, and the training data is represented as
\begin{equation}
\label{equ:dataset}
\bigcup\limits_{t=1}^T \bigcup\limits_{i=1}^N (P_{t,i}, \Delta \theta_{t,i}),
\end{equation}
where $T$ is the number of sampling period, $N$ is the number of images in raw dataset.

The sampling process is illustrated in Figure~\ref{fig:Framework}(a). By the sampling process, we obtained nearly unlimited training data and the training space contains most possible landmarks' distribution from coarse stages to fine stages. The sampled training data is online generated to save the system memory.
\begin{algorithm}[ht]
\begin{algorithmic}[1]
\REQUIRE Raw face landmarks dataset: $\bigcup\limits_{i=1}^N (I_{i}, (S_{gt})_{i})$
\ENSURE Training dataset: $\bigcup\limits_{t=1}^T \bigcup\limits_{i=1}^N (P_{t,i}, \Delta \theta_{t,i})$
\FOR {$t = 1$ to $T$}
    \FOR{$i = 1$ to $N$}
        \STATE random select value $S_{t, i}$ in Equ.~\eqref{equ:sampling_space};
        \STATE Calculate location $\theta_{t, i}$ by $S_{t, i}$ by Equ.~\eqref{equ:face_model};
        \STATE Extract patches $P_{t, i}$ for image $I_i$ in location $\theta_{t, i}$;
        \STATE Set $P_{t, i}$ as the regressor's input data;
        \STATE Set $\Delta \theta_{t, i} = \theta_i^* - \theta_{t, i} $ as regressor's target increment;
    \ENDFOR
\ENDFOR
\end{algorithmic}
\caption{Sampling process of SIR}
\label{alg:sampling}
\end{algorithm}

\subsection{Landmarks-Attention Network}
In this section, we will describe the structure of the proposed regressor. Our goal is to learn a mapping between appearance features and landmarks' location increment. Previous works usually first obtain robust initialization location by extracting features in the whole image and then refine the location by many refinement networks~\cite{XiaoFXLYK:RAR:ECCV2016,SunWT:cascaded-cnn:cvpr13} or stack all landmarks patches to directly extract all landmarks features~\cite{Trigeorgis:MDM:CVPR16}. They either require a number of model parameters or generate indiscriminative features. Thus we propose a Landmarks-Attention Network (LAN) to overcome the above two drawbacks. Our regressor is a single CNN which concurrently \emph{pays attention to} appearance feature around each facial landmark. Specifically, for each landmarks patch, we extract features by several convolutional and pooling layers, then concatenate these independent feature vectors and add two fully connected layers to learn a holistic location increment. The structure of each feature extraction sub-network is illustrated in Figure~\ref{fig:Framework}(b) and the detailed information of the sub-network is presented in Table~\ref{tab:SubNetwork}.
\begin{table}[h]
\centering
\caption {Feature extraction sub-network of Landmarks-Attention Network for each patch.}
\label{tab:SubNetwork}
\scriptsize
\begin{tabular}{|c|c|c|c|}
\hline
Layer & Input Tensor & Kernel & Output Tensor\\
\hline
conv1 & $57\times57\times3$ & $3\times3\times3\times16$ & $57\times57\times16$ \\
\hline
pool1 & $57\times57\times16$ & $2\times2$ & $29\times29\times16$ \\
\hline
conv2 & $29\times29\times16$ & $2\times2\times16\times32$ & $29\times29\times32$ \\
\hline
pool2 & $29\times29\times32$ & $2\times2$ & $15\times15\times32$ \\
\hline
conv3 & $15\times15\times32$ & $2\times2\times32\times64$& $15\times15\times64$ \\
\hline
pool3 & $15\times15\times64$ & $2\times2$& $8\times8\times64$ \\
\hline
fc1 & \makecell{$8\times8\times64$ \\ $(1\times1\times4096)$} & $4096\times10$ & $1\times1\times10$\\
\hline
\end{tabular}
\end{table}

Compared to the previous networks, our proposed model has three advantages: (1) The landmarks feature extracted by independent sub-networks can be more discriminative, as showed in Figure~\ref{fig:network_curve}; (2) Concatenating all independent features vectors and adding fully connected layers can obtain a holistic landmarks location increment, especially when some landmarks are occluded or blurred; (3) Our network is very light, whose parameters number(3.72M in total) is far less than other CNN models (\eg, AlexNet~\cite{alexnet} contains about 60M parameters and VGGNet~\cite{VGG} contains about 138M parameters).

\subsection{Training}
The training process is illustrated in Figure~\ref{fig:Framework} (a) and (b). Since sampling period $T$ can be large enough, online random sampling process can generate nearly unlimited training data $\bigcup\limits_{t=1}^T \bigcup\limits_{i=1}^N (P_{t,i}, \Delta \theta_{t,i})$. Then, the above described LAN is trained to learn the descent directions in coarse and fine stages together. This process can be formulated as
\begin{equation}
R_{\Delta} = \underset{R_\Delta}{\operatorname{argmin}}\frac{1}{T\times N}\sum_{t=1}^{T}\sum_{i=1}^{N}\|\Delta\theta_{t,i}-R_{\Delta}(P_{t,i}) \|_2^2,
\end{equation}
where $R_\Delta$ is the target self-iterative regressor (\ie, LAN), and $t$ indicates the $t_{th}$ sampling period.

\begin{figure}[ht]
\begin{subfigure}{.25\textwidth}
  \centering
  \includegraphics[width=\linewidth]{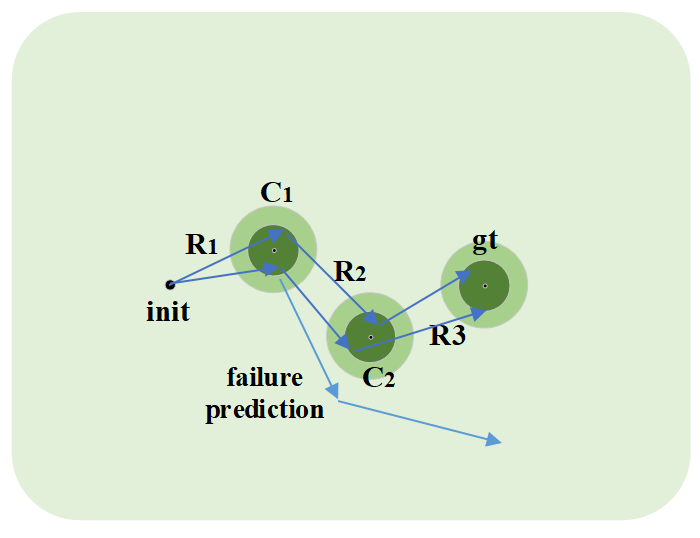}
  \caption{Cascaded Regression}
  \label{fig:sample_fig1}
\end{subfigure}%
\begin{subfigure}{.25\textwidth}
  \centering
  \includegraphics[width=\linewidth]{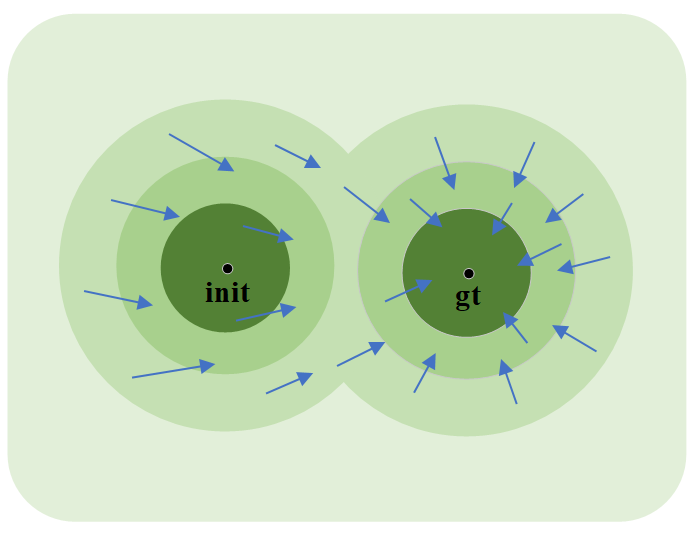}
  \caption{Self-Iterative Regression}
  \label{fig:sample_fig2}
\end{subfigure}
\caption{(a) Typical cascaded regression process: starting from initial value, parameters are updated and close to the ground truth (such as $init\rightarrow C_1\rightarrow C_2\rightarrow gt$) by regressors $R_k(k=1,2,3,...)$. Once one regressor predicts the false direction, the final result is prone to drift away; (b) SIR Descent Direction Map: the training space of SIR includes distribution from coarse stages to fine stages and all descent directions are pointed to ground truth. }
\label{fig:sampling_space}
\end{figure}

Since the training space of SIR includes most possible landmarks distribution from coarse stages to fine stages, the training process will generate a Descent Direction Map (DDM) in the sampling space where each sample's descent direction roughly points toward the ground truth. As illustrated in Figure~\ref{fig:sampling_space} (b), SIR is more robust than CR because the former can cover more training space and isn't affected by the optimization path. When the previous regressor predicts false descent directions, SIR can still converge to the ground truth while CR is prone to drift away.

% \begin{figure}[ht]
% \centering
% \includegraphics[width=0.95\linewidth]{Figs/Sample_Space_SIR_VS_Cascaded.PNG}
% \caption{\small (a) SIR descent direction map: the training space of SIR includes distribution from coarse stages to fine stages and all descent directions are pointed to ground truth; (b) Typical cascaded regression process: starting from initial value, parameters are updated and close to the ground truth (such as \text{init}$\rightarrow$\text{C1}$\rightarrow$\text{C2}$\rightarrow$\text{gt}) based on regressors $R_k(k=1,2,...K)$. Once one regressor predicts the false direction, the final result is prone to drift away.}
% \label{fig:sampling_space}
% \end{figure}

\subsection{Self-Iterative Updating}
For the testing process, similar to the cascaded regression methods, starting from initial location parameters $\theta_0$, we iteratively update the location parameters $\theta_k$ and extract new patches $P_k$ till converges. The process is presented in Algorithm~\ref{alg:testing}, and facial landmarks location parameter is updated by, 
\begin{equation}
    \theta_{k+1} = \theta_{k} + R_{\Delta}(P_k), \quad k = 0, 1,\cdots.
\label{equ:iterative-updating}
\end{equation}

\begin{algorithm}[h]
\begin{algorithmic}[1]
\REQUIRE Regressor $R_{\Delta}$, Initial location $\theta_{0}$, Total iteration times $K$
\ENSURE Prediction of facial landmarks' location $\theta_{K}$
\FOR {$k = 0$ to $K-1$}
\STATE Extract patches $P_k$ in location $\theta_k$\;
\STATE $\theta_{k+1} = \theta_k + R_{\Delta}(P_k)$\;
\ENDFOR
\end{algorithmic}
\caption{Self-Iterative updating process of SIR}
\label{alg:testing}
\end{algorithm}

\section{Experiments}\label{sec:experiment}
In this section, we perform experiments to demonstrate the effectiveness of the proposed SIR compared to state-of-the-art methods. Specifically, we evaluate the proposed method model by (1) comparing the performance of SIR vs. state-of-the-art and baseline cascaded regression; (2) comparing the number of model parameters and memory storage of pre-train models; and (3) studying the effect of the proposed feature extraction network(LAN), the number of iteration times and sampling space parameter.

{\noindent \textbf{Datasets.}} The 300-W dataset is short for $300$ faces in-the-wild~\cite{300W}, which is designed for evaluating the performance of facial landmarks detection. The training set ($3,148$ faces in total) consists of AFW dataset~\cite{Ramanan:AFW:2012}, HELEN training set~\cite{Helen:2012} and LFPW training set~\cite{Belhumeur:2011:LFPW}. Two testing sets are established, \ie, \textit{public testing set} ($689$ faces in total) including HELEN testing set~\cite{Helen:2012}, LFPW testing set~\cite{Belhumeur:2011:LFPW} and IBUG dataset~\cite{300W}; and \textit{competition testing set} ($600$ faces in total) including $300$ indoor and $300$ outdoor faces images.

{\noindent \textbf{Metrics.}} Normalized Mean Error (NME) measures landmarks' mean location error normalized by inter-pupil (eyes centers) distance~\cite{Zhu:CFSS:CVPR2015,ShaoqinRen:LBF:CVPR2014} or inter-ocular (outer eye corners) distance~\cite{Trigeorgis:MDM:CVPR16}. Cumulative Error Distribution (CED) curve is the cumulative distribution function of the normalized error, which can avoid heavily impacted by some big failures~\cite{survey:corr:YangJLR15}. We also calculated another two evaluation metrics, namely Area-Under-the-Curve (AUC$_{\alpha}$) and Failure Rate (FR$_{\alpha}$). Similar as MDM~\cite{Trigeorgis:MDM:CVPR16}, we consider mean point-to-point error greater than $0.08$ as a failure, \ie, $\alpha=0.08$.

{\noindent \textbf{Implementation Detail.}} We perform the experiments based on a machine with Core i7-5930k CPU, 32 GB memory and GTX 1080 GPU with 8G video memory. The detected faces are resized into $256 \times 256$ and the location patch size is $57 \times 57$. For CNN structure, the Rectified Linear Unit (ReLU) is adopted as the activation function, and the optimizer is the Adadelta~\cite{adadelta} approach, learning rate is set to $0.1$ and weight decay is set to $1e-4$. Training the CNN requires around $2$ days.

\subsection{Comparison with State-of-the-arts}
\begin{figure}[h]
\centering
\includegraphics[width=.85\linewidth]{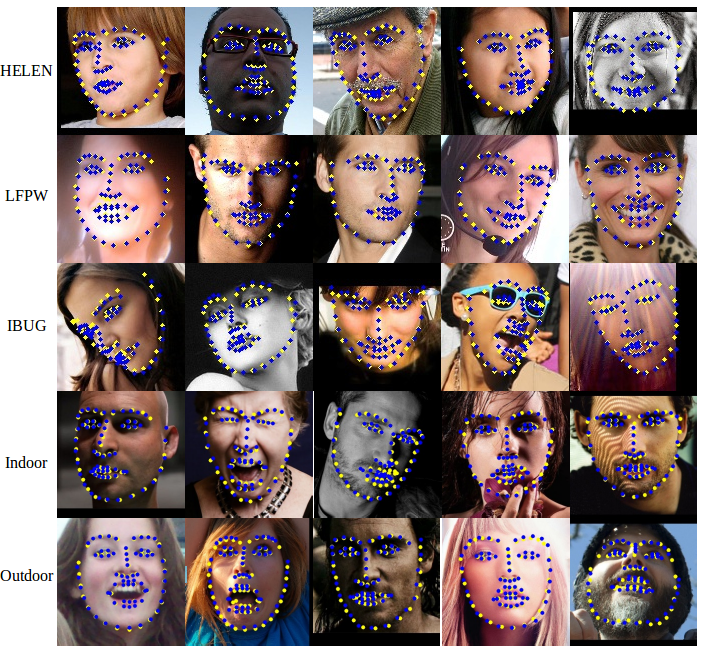}
\caption{Several facial landmarks detection results in 300-W public testing and competition testing set. Blue dot in each sub-picture indicates ground truth landmarks location and yellow dot indicates the predicted location of SIR. Pictures for the five rows are from HELEN testing set, LFPW testing set, IBUG set, 300-W competition testing Indoor and Outdoor set respectively.}
\label{fig:visualization}
\end{figure}

As shown in Table~\ref{tab:nme:public}, we compare the proposed method with several state-of-the-art facial landmarks detection methods in the  public testing set. Specifically, the \textit{common subset} consists of LFPW testing set (224 faces) and HELEN testing set (330 faces) and the \textit{challenging subset} is IBUG dataset (135 faces). Thus the the \textit{full set} (689 faces) of the union of the common (554 faces) and challenging subsets (135 faces). The NME results shows that SIR performs comparatively with RAR~\cite{XiaoFXLYK:RAR:ECCV2016} and outperform other existing methods~\cite{Cao:ESR:CVPR2012,Burgos:RCPR:ICCV2013,Xiong:SDM:CVPR2013,ShaoqinRen:LBF:CVPR2014,Zhu:CFSS:CVPR2015,KowalskiN:K-Regression:SPL16,Trigeorgis:MDM:CVPR16,XiaoFXLYK:RAR:ECCV2016}. Besides, more visual results are also illustrated in Figure~\ref{fig:visualization}. In the more challenging IBUG subset, our method achieves robust performance in large pose, expression and illumination environment.
% table 
\begin{table}[h]
\centering
\caption {NME (inter-pupil normalization) results in the  public testing set. The top two performance are shown in boldface.}
\label{tab:nme:public}
\centering
\setlength{\tabcolsep}{3.5pt}
\footnotesize{
\begin{tabular}{c c c c}
\hline
Methods &\makecell{Common\\subset} &\makecell{Challenging\\subset} &\makecell{Full\\set} \\
\hline
% RCPR~\cite{Burgos:RCPR:ICCV2013} & 6.18 & 17.26 & 8.35 \\
% ESR~\cite{Cao:ESR:CVPR2012} & 5.28 & 17.00 & 7.58 \\
% SDM~\cite{Xiong:SDM:CVPR2013} & 5.57 & 15.40 & 7.50 \\
% LBF~\cite{ShaoqinRen:LBF:CVPR2014} & 4.95 & 11.98 & 6.32\\
% CFAN~\cite{eccv14:ZhangSKC:CFAN} & 5.55 & - & - \\
% CFSS~\cite{Zhu:CFSS:CVPR2015} & 4.73 & 9.98 & 5.76 \\
% Kowalski~\etal~\cite{KowalskiN:K-Regression:SPL16} & 4.62 & 9.48 & 5.57 \\
% MDM~\cite{Trigeorgis:MDM:CVPR16} & 4.83 & 10.14 & 5.88\\
% RAR~\cite{XiaoFXLYK:RAR:ECCV2016} & \textbf{4.12} & \textbf{8.35} & \textbf{4.94}\\

RCPR(2013) & 6.18 & 17.26 & 8.35 \\
ESR(2012)& 5.28 & 17.00 & 7.58 \\
SDM(2013) & 5.57 & 15.40 & 7.50 \\
LBF(2014) & 4.95 & 11.98 & 6.32\\
CFAN(2014) & 5.55 & - & - \\
CFSS(2015) & 4.73 & 9.98 & 5.76 \\
Kowalski~\etal(2016) & 4.62 & 9.48 & 5.57 \\
MDM(2016) & 4.83 & 10.14 & 5.88\\
RAR(2016) & \textbf{4.12} & \textbf{8.35} & \textbf{4.94}\\
\hline
\textbf{SIR} & \textbf{4.29} & \textbf{8.14} & \textbf{5.04} \\
\hline
\end{tabular}}
\end{table}

\begin{figure*}[ht]
  \begin{subfigure}[b]{0.35\textwidth}
    \includegraphics[width=\textwidth]{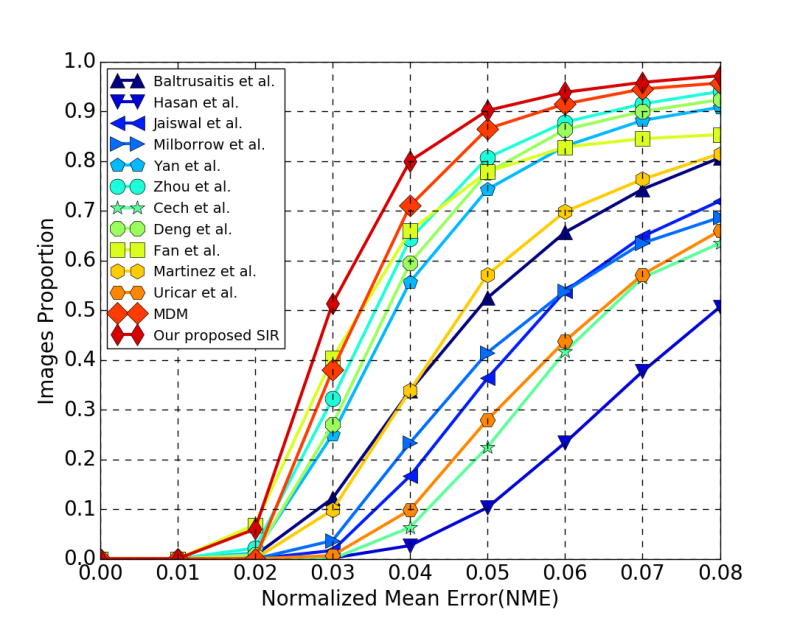}
    \caption{CED curve of facial \emph{51-points}.}
    \label{fig:f1}
  \end{subfigure}
  \qquad \qquad \qquad \qquad
  \begin{subfigure}[b]{0.35\textwidth}
    \includegraphics[width=\textwidth]{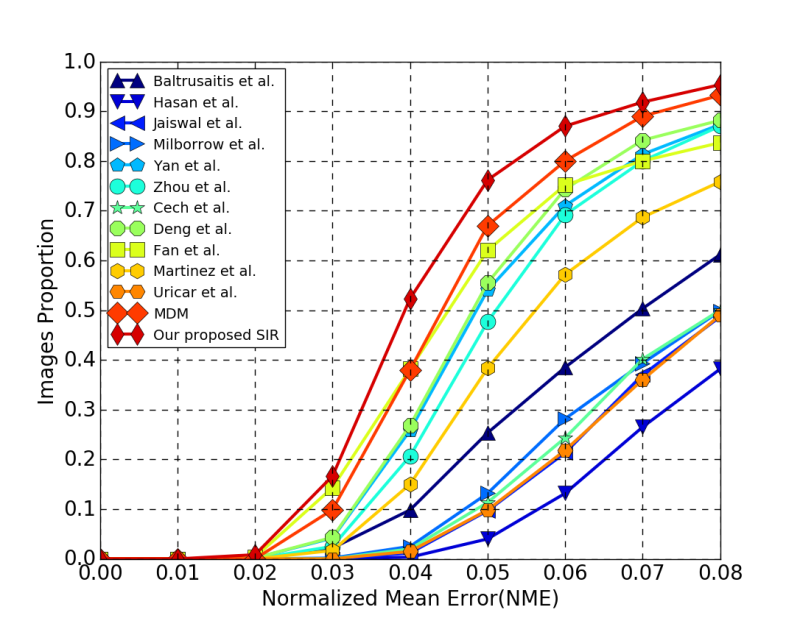}
    \caption{CED curve of facial \emph{68-points}.}
    \label{fig:f2}
  \end{subfigure}
  \caption{CED curve results comparison in 300-W competition testing set.}
\label{fig:CED}
\end{figure*}

% end table
On the other hand, we evaluate SIR in the  competition testing set. As shown in Figure~\ref{fig:CED}, the SIR method outperform the state-of-the-art methods~\cite{Cech:2016:MFL,Deng:2016,Fan:2016:AHL,Baltrusaitis:2013,Yan:2013,Zhou:Face++:2013,ivc:UricarFTSH16,Jaiswal_2013_ICCV_Workshops,Milborrow_2013_ICCV_Workshops,Hasan_2013_ICCV_Workshops,Martinez:2016:LRP} according to the CED curve. Moreover, Table~\ref{tab:auc:failure} presents the quantitative results for both the 51-point and 68-point error metrics (\ie, AUC and Failure Rate at a threshold of $0.08$ of the normalised error), compared to existing methods~\cite{KazemiS:ERT:CVPR2014,Tzimiropoulos:PO-CR:CVP2015,AsthanaZCP:Chehra:CVPR2014,Xiong:SDM:CVPR2013,Zhou:Face++:2013,Yan:2013,ivc:UricarFTSH16,Zhu:CFSS:CVPR2015,Trigeorgis:MDM:CVPR16}. The promising performances on two metrics indicate the effectiveness of the proposed method.

\begin{table}[h]
\centering
\caption{Quantitative results using AUC$_{0.08}$(\%) and FR$_{0.08}$(\%) in the  competition testing set. \emph{51-points} and \emph{68-points} are two groups of facial landmarks and \emph{51-points} is part of \emph{68-points}.}
\setlength{\tabcolsep}{8pt}
\label{tab:auc:failure}
\centering
\footnotesize{
  \begin{tabular}{l|l l|l l}
    {} &
      \multicolumn{2}{c|}{\textbf{51-points}} & \multicolumn{2}{c}{\textbf{68-points}} \\
    \hline
    \textbf{Method} & AUC & Failure & AUC & Failure \\
    \hline
    % ERT~\cite{KazemiS:ERT:CVPR2014} & 40.60 & 13.50 & 32.35 & 17.00 \\
    % PO-CR~\cite{Tzimiropoulos:PO-CR:CVP2015} & 47.65 & 11.70 & - & - \\
    % Chehra~\cite{AsthanaZCP:Chehra:CVPR2014} & 31.12 & 39.30 & - & - \\
    % Intraface~\cite{Xiong:SDM:CVPR2013} & 38.47 & 19.70 & - & - \\ 
    % Balt \etal~\cite{Baltrusaitis:2013} & 37.65 & 17.17 & 19.55 & 38.83 \\
    % zhou \etal~\cite{Zhou:Face++:2013} & 53.29 & 5.33 & 32.81 & 13.00 \\
    % Yan \etal~\cite{Yan:2013} & 49.07 & 8.33 & 34.97 & 12.67 \\
    % CFSS~\cite{Zhu:CFSS:CVPR2015} & 50.79 & 7.80 & 39.81 & 12.30 \\
    % MDM~\cite{Trigeorgis:MDM:CVPR16} & 56.34 & 4.20 & 45.32 & 6.80 \\
    ERT(2014) & 40.60 & 13.50 & 32.35 & 17.00 \\
    PO-CR(2015) & 47.65 & 11.70 & - & - \\
    Chehra(2014) & 31.12 & 39.30 & - & - \\
    Intraface(2013) & 38.47 & 19.70 & - & - \\ 
    Balt \etal(2013) & 37.65 & 17.17 & 19.55 & 38.83 \\
    zhou \etal(2013) & 53.29 & 5.33 & 32.81 & 13.00 \\
    Yan \etal(2013) & 49.07 & 8.33 & 34.97 & 12.67 \\
    CFSS(2015) & 50.79 & 7.80 & 39.81 & 12.30 \\
    MDM(2016) & 56.34 & 4.20 & 45.32 & 6.80 \\
    \hline
    \textbf{SIR} & \textbf{58.11} & \textbf{2.83} & \textbf{46.56} & \textbf{4.33} \\
    \hline
  \end{tabular}}
\end{table}

\subsection{Comparison with Cascaded Regression}
As discussed before, previous cascaded regression methods adding more regressors can achieve better performance, but increase the number of model parameters, computing resources and storage space, especially for deep learning based methods. Different from them, our method obtains state-of-the-art performance by iterative call the same regressor rather than adding any more regressors.

Our method reduces the model complexity while keeps the performance in two folds: (1) the proposed network focuses on the landmarks' local feature, which significantly reduces the dimension of final feature layer; (2) only one CNN module is required to iteratively predict the location parameters, while cascaded regression usually requires at least three regressors\cite{Trigeorgis:MDM:CVPR16,Xiong:SDM:CVPR2013}. 

To prove the effectiveness of SIR, we add a baseline CR method which extracts features by the same LAN while adopts cascaded regression framework. Both baseline CR and SIR is updated for 4 times before the stable performance. As shown in Table~\ref{tab:parameters}, our method requires parameters and memory far less than other cascaded regression based methods.
% On the other hand, it is worth mentioning that adding more regressors in our framework will slightly decrease the performance. \red{This is because XXX.}

\begin{table}[h]
\centering
\caption {Comparison with state-of-the-art methods in the  public testing set, with the first and the second best results highlighted. DL indicates whether the method is based on deep learning.}
\label{tab:parameters}
\footnotesize{
\begin{tabular}{c c c c c}
\hline
Method & DL & NME & \# params & model memory\\
\hline
% RCPR~\cite{Burgos:RCPR:ICCV2013} & N & 8.35 & - & 91.3MB\\
% LBF~\cite{ShaoqinRen:LBF:CVPR2014} & N & 6.32& - & 36.6MB \\
% CFSS~\cite{Zhu:CFSS:CVPR2015} & N & 5.76 & - & 225.2MB\\
% MDM~\cite{Trigeorgis:MDM:CVPR16} & Y & 5.61 & 80.00M & 322.3MB \\
% RAR~\cite{XiaoFXLYK:RAR:ECCV2016} & Y & \textbf{4.94} & 14.58M+  & 58.3MB+\\ 
RCPR(2013) & N & 8.35 & - & 91.3MB\\
LBF(2014) & N & 6.32& - & 36.6MB \\
CFSS(2015) & N & 5.76 & - & 225.2MB\\
MDM(2016) & Y & 5.61 & 80.56M & 322.3MB \\
RAR(2016) & Y & \textbf{4.94} & 15.65M+  & 62.6MB+\\
\hline
\textbf{baseline CR} & Y &  6.23 & \textbf{14.88M} & \textbf{62.4MB} \\
\textbf{SIR} & Y & \textbf{5.04} & \textbf{3.72M} & \textbf{15.6MB}\\
\hline
\end{tabular}}
\end{table}

\subsection{Discussion and Analysis}\label{sec:discussion}
In this section, we perform analyses on the effect of several important modules in our method to the final performance.

\begin{figure}[H]
\centering
\includegraphics[width=0.75\linewidth]{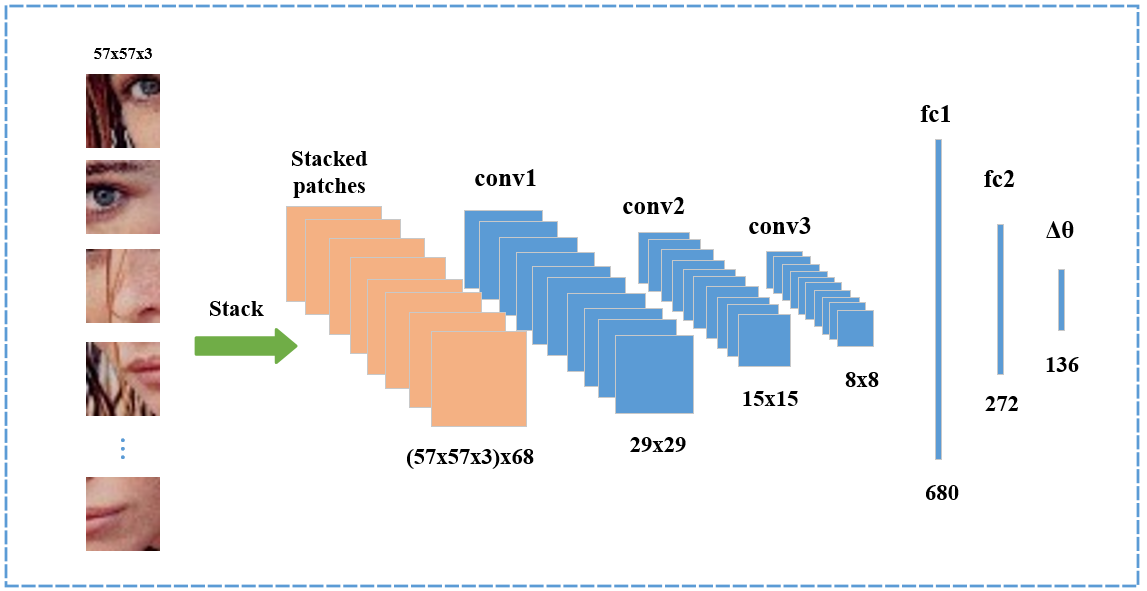}
\caption{Structure of SIR-Stack Network. For a fair comparison, we adopt the same number of CNN layers and model parameters(3.72M).}
\label{fig:stack_patch}
\end{figure}

\begin{figure}[h]
\centering
\includegraphics[width=0.75\linewidth]{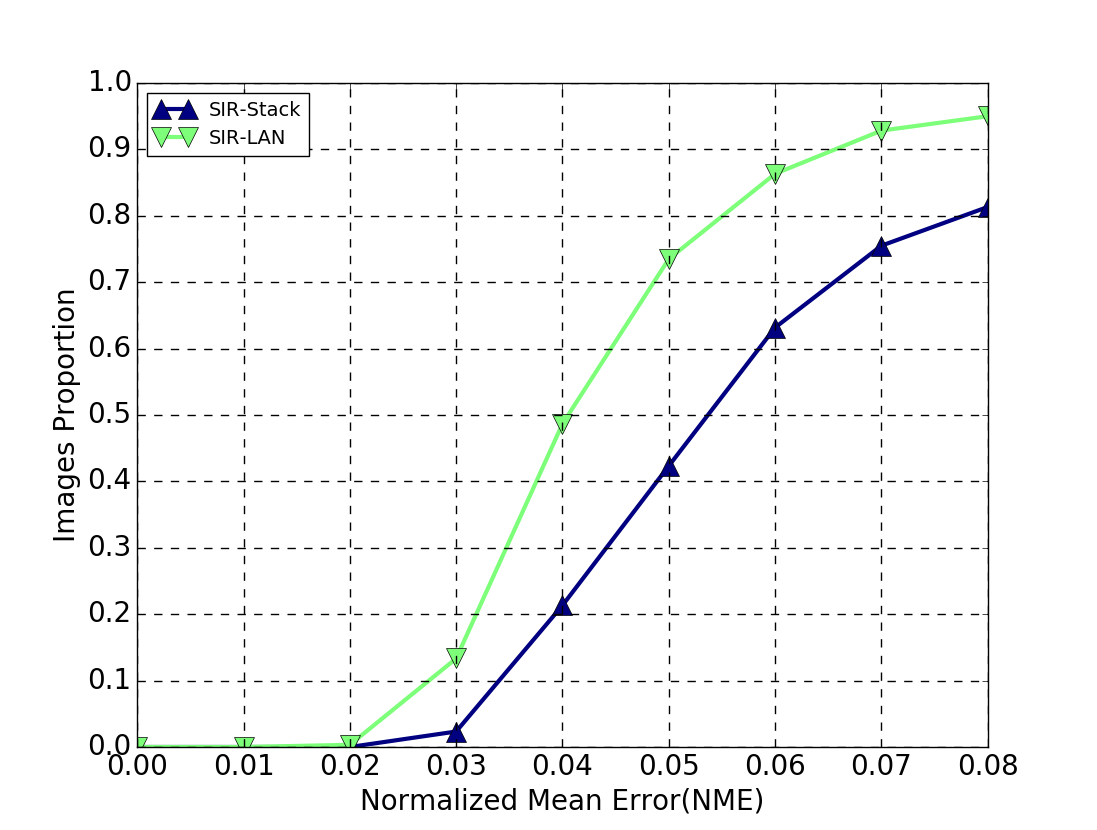}
\caption{Comparison between SIR-LAN and SIR-Stack in the competition testing set.}
\label{fig:network_curve}
\end{figure}

{\noindent \textbf{Effect of different feature extraction networks.}} In SIR framework, we adopt the Landmarks-Attention Network (we call it SIR-LAN) to extract landmarks patches features separately, while some works stack all landmarks patches and then extract the whole features directly (we call it SIR-Stack), as illustrated in Figure~\ref{fig:stack_patch}. To demonstrate the effectiveness of our network, we conduct an experiment by SIR framework to compare the above two networks with the same number of CNN layers and model parameters, the structure of SIR-Stack is showed in Figure~\ref{fig:stack_patch}. The result illustrated in Figure~\ref{fig:network_curve} shows that the proposed network extracting patches features separately performs significantly better than previous methods extracting patches feature together (\eg, MDM~\cite{Trigeorgis:MDM:CVPR16}).

\begin{figure}[H]
\centering
\includegraphics[width=0.80\linewidth]{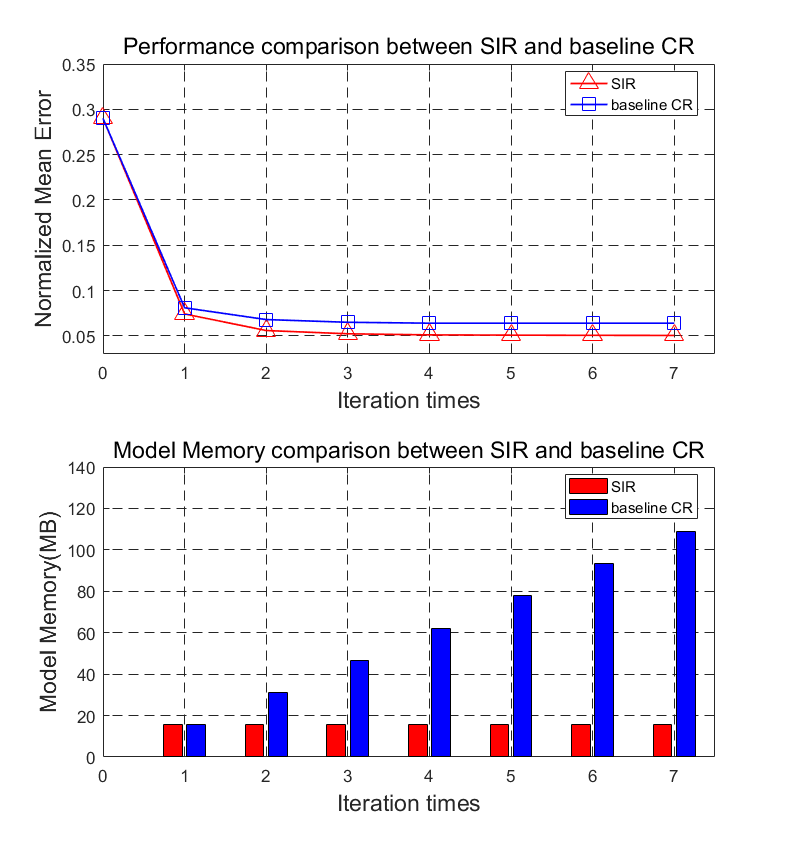}
\caption{Effect of iteration times. Top: Comparison between SIR and baseline CR in accuracy. With the increase of iteration times, both SIR and baseline CR can decrease the detection error and SIR performs better than baseline CR. Bottom: Comparison between SIR and baseline CR in in Model Memory. Increasing the iteration times will increase its model memory of baseline CR, while SIR doesn't because it can iteratively call itself.}
\label{fig:iteration_curve}
\end{figure}

{\noindent \textbf{Effect of iteration times.}} From Figure~\ref{fig:iteration_curve}, we can find that the accuracy will be improved by adding iteration times before the stable performance (\ie, $4$ iterations) is achieved. When increasing iteration times, more model memory will be added in baseline CR.

\begin{figure}[H]
\centering
\includegraphics[width=0.75\linewidth]{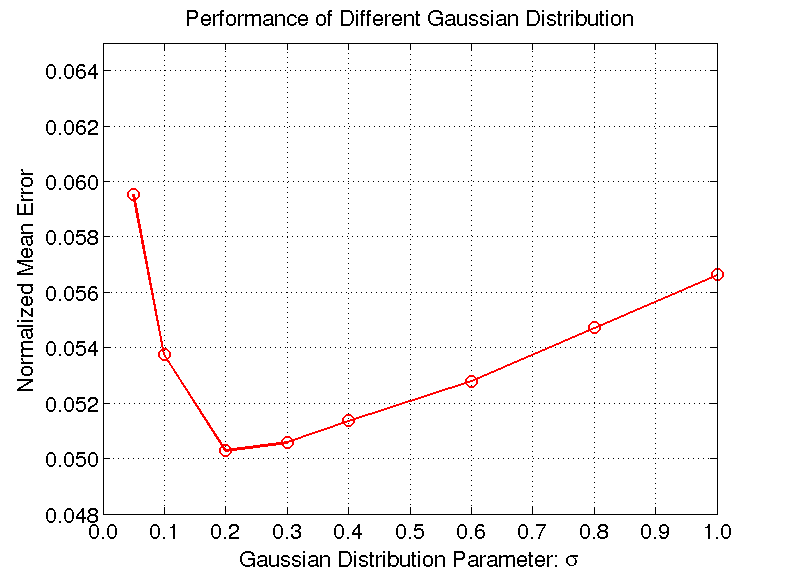}
\caption{Performances of different Gaussian sampling in the 300-W public testing set.}
\label{fig:sigma_nme_curve}
\end{figure}

{\noindent \textbf{Effect of Gaussian sampling space parameters.}} As one of the most important processes, random sampling space significantly affects the final robustness and accuracy. As shown in Figure~\ref{fig:sigma_nme_curve}, the NME results are presented by varying the standard deviation $\sigma$ of Gaussian sampling. Appropriate values lead to promising performance so that we set $\sigma=0.2$ in our method.

\section{Conclusion}
In this paper, we develop a SIR framework solve the non-linear least squares problems. Compared with cascaded regression, it only needs to train a single regressor to learn descent directions in coarse stages to fine stages together, and refines the target parameters iteratively by call the same regressor. Experimental results in the facial landmarks detection task demonstrate that the proposed self-iterative regressor achieves comparable accuracy to state-of-the-art methods, but significantly reduces the number of parameters and memory storage of the pre-trained models. In the future, we will extend the proposed method to other applications, such as human pose prediction, structure from motion and 3D face reconstruction.

{\small
\bibliography{ref}
\bibliographystyle{aaai}
}
\end{document}